# Hyperspectral Unmixing using a Structured Linear Mixing Model with Additive Spectral Variability

Salah Eddine Brezini, Yannick Deville, Moussa Sofiane Karoui, Fatima Zohra Benhalouche and Abdelaziz Ouamri

*Abstract*— **Linear hyperspectral unmixing aims at retrieving pure material spectra present in an observed scene and their associated abundance fractions in each pixel. In most practical situations, hyperspectral images are affected by spectral variability (SV) also known as intra-class variability, which is caused by different physical phenomena including atmospheric effects, intrinsic variability of the pure materials and illumination or topographic changes. Nevertheless, the standard hyperspectral unmixing methods that rely on the Linear Mixing Model (LMM) do not consider these phenomena. Thus, neglecting them will spread errors throughout the linear hyperspectral unmixing process. To avoid this problem, this paper proposes two hyperspectral unmixing algorithms dedicated to a parametric model called Structured Additive Variability LMM (SAV-LMM). This mixing model handles SV in an additive manner by incorporating two terms into the standard LMM. Moreover, the designed algorithms are based on Nonnegative Matrix Factorization (NMF) and optimize a cost function which integrates additional penalization accounting for SV. Furthermore, a novel improved initialization scheme, which takes SV into account, is proposed. Experiments conducted on both synthetic and real datasets show the effectiveness of the proposed algorithms for linearly unmixing hyperspectral images with SV compared to the tested state-of-the-art methods.**

## I. INTRODUCTION

Remote sensing hyperspectral images exhibit a fine spectral resolution since they are acquired by sensors that collect light scattered by objects present in its field of view in hundreds of narrow and contiguous spectral bands covering the visible-infrared domain [1]. Thus, this rich spectral information enables an accurate identification of pure materials present in an observed hyperspectral image (HSI). This feature allows HSIs to be at the core of different space applications [2] including environmental monitoring [3], sustainable energy [4], and that undoubtedly improves our knowledge about the Earth surface and the physical phenomena surrounding us.

Salah Eddine Brezini is with the Université des Sciences et de la Technologie d'Oran, Oran, Algeria and with Univ Toulouse, CNES, CNRS, IRAP, Toulouse, France (e-mail: salaheddine.brezini@univ-usto.dz).

Yannick Deville is with Univ Toulouse, CNES, CNRS, IRAP, Toulouse, France (e-mail: yannick.deville@utoulouse.fr)

Moussa Sofiane Karoui and Fatima Zohra Benhalouche are with the Centre des Techniques Spatiales, Agence Spatiale Algérienne, Arzew, Algeria, and with Univ Toulouse, CNES, CNRS, IRAP, Toulouse, France (e-mails: {moussa-sofiane.karoui, fatima-zohra.benhalouche}@utoulouse.fr).

Abdelaziz Ouamri is with the Université des Sciences et de la Technologie d'Oran, Oran, Algeria (e-mail: ouamri@yahoo.com).

Nevertheless, since the acquisition of HSIs is achieved in narrow bandwidths, the remote hyperspectral sensors do not obtain enough photons to maintain a high signal-to-noise ratio (SNR) [5], [6]. This physical limitation compels the hyperspectral sensors to work by operating an optimal compromise between satisfactory SNR and spatial resolution [5]. This physical constraint inevitably provides an HSI with low spatial resolution and leads to the well-known problem of the existence of mixed pixels in HSIs [7]. Hence, it hinders their applications in many fields which need both high spectral and spatial resolutions. In this case, each observed pixel spectrum is represented by the mixture of spectral contributions from pure materials (also called endmembers), present in the considered HSI, and their associated abundance coefficients. The way the spectral combination is achieved depends directly on the observed scene [6]. To circumvent this physical limitation, the Spectral Unmixing (SU) remains relevant. SU techniques aim to estimate a collection of pure material spectra and their associated abundance coefficients. SU can be seen as a typical Blind Source Separation (BSS) problem [8], [9]. Nowadays, various methods, based on the Linear Mixture Model (LMM) [1], are being developed. Due to their simplicity, they are primarily known as Linear Spectral Unmixing (LSU) techniques. Indeed, LMM is the result of the basic assumption, which states that an endmember can be represented by only one spectrum in the whole HSI and the latter can be fully reconstructed from this assumption. However, this assumption does not hold anymore when some physical phenomena occur in the observed scene. Consequently, LMM appears rather limited by two main issues [10], [11], namely the spectral/intra-class variability [10], [12], [13] and the nonlinearity [11], [14]-[16].

The present paper aims to address the Spectral Variability (SV) issue. It is often caused by several factors such as: illumination and topographic changes, atmospheric conditions and/or material composition [10], [12], [13]. To tackle SV, the concept of endmembers is substituted by the notion of classes of endmembers. To this end, various methods have been developed. Techniques tackling SV can be grouped into three classes [10], [12]: (1) spectral library-based approaches, (2) methods using a particular statistical distribution to describe observed data, and (3) parametric model-based methods. A straightforward manner to tackle SV is to consider spectral libraries. Indeed, they can be obtained using two approaches: the first one is to gather spectra obtained from a field campaign [17] and the second one is to extract endmember spectra directly from the considered HSI [18]-[21] by running an Endmember Extraction Algorithm (EEA) several times on the considered HSI.

Recent approaches handle SV by representing an HSI as a statistical distribution [22]-[25]. Furthermore, parametric model-based techniques [10] integrate SV directly into LMM by considering variables representing SV. This class of methods is more flexible since neither a spectral library nor prior knowledge of the statistical distribution of the involved variables is needed to achieve LSU. Hence, different methods have recently been introduced [26]-[31]. In [26], the perturbed LMM (PLMM) has been proposed to handle SV by adding perturbation terms to endmembers. PLMM jointly estimates endmember spectra and abundance fractions. The extended LMM (ELMM) [27] represents SV arising from illumination and topography changes in each pixel by using scaling factors. Recently, a generalization of ELMM, termed as the generalized LMM (GLMM) [28], has been introduced. Among its advantages, GLMM allows the scaling factors to vary within each spectral band, providing GLMM with more flexibility with respect to ELMM. The augmented LMM (ALMM) [29] adds low-rank terms to ELMM to overcome its limitation by addressing more complex SV that cannot just be described by illumination changes. Lately, the focus of recent research has been on reducing the interdependencies among the abundance estimations [32]. In [33], the authors use spatial contextual information to improve the estimation of abundance fractions. In [34], a sparse representation is used to tackle SV.

The above methods generally perform well in the case of SV induced by illumination and topography changes, particularly the parametric model-based methods. In fact, they do not explicitly take into account the intrinsic variability caused by the chemical composition of a material, which is more complex to treat and cannot be just represented by scaling factors. Indeed, these methods deal with SV mainly in the spatial part of variables and do not consider it in the spectral side of the involved variables. To this end, new sophisticated LMMs have been developed that allow addressing SV in both spectral and spatial domains [35]. Furthermore, this extended version of LMM, which explicitly models SV not only in the spatial domain but also in the spectral domain, makes it possible to retrieve each pixel of an HSI the associated spectra for each class of endmembers. The first technique using this concept is known as Inertia-Constrained Pixel-by-Pixel NMF (IP-NMF) [35]. Furthermore, the assumption which states that SV is not only induced by topographic or/and illumination changes is supported by a data analysis in [35], which shows that in urban scenes, the intrinsic variability is present and cannot be neglected in the unmixing process. To overcome this shortcoming, IP-NMF penalizes the trace of the estimated spectra covariance matrix for each class to tune the inertia of each class of endmembers. IP-NMF proves to be very efficient in urban environments and for ocean remote sensing [3], [36].

In [37]-[39], a novel parametric model has been introduced, without any physical consideration, by adding two matrices to the LMM to address SV. Indeed, this variation can be defined as tunable additive shifts around an adaptive reference spectrum per endmember for each class of endmembers. Moreover, parametric models in [38] and [39] consider structured variables to deal more efficiently with SV.

In this paper, a complete flowchart is proposed by considering SV from the initialization stage to the estimation of endmember spectra and abundance fractions. To this end, two algorithms, based on Nonnegative Matrix Factorization (NMF) [40], [41], dedicated to a proposed parametric mixing model that deals with SV, are introduced and an improved initialization approach is proposed. More precisely, the designed algorithms are gradient descent-based ones. The first one, called Gradient-based Structured Additive Variability LMM (Grad-SAV-LMM), is based on the gradient descent algorithm [42]. The second one, termed Multiplicative SAV-LMM (Multi-SAV-LMM), considers the multiplicative gradient-based update rules [43].

As a matter of fact, the proposed methods and the considered state-of-the-art techniques address ill-posed optimization problems and the performance of the derived algorithms is directly affected by two important factors, which are noise present in the image and the initialization. To improve their performance, a new strategy is introduced by using a (pseudo or image-based) spectral library, obtained by means of the Automated Extraction of image-based Endmember Bundles (AEEB) algorithm [18], into the initialization stage.

The main contributions of our article can be outlined as follows:

1) Deal with SV by allowing more spectral flexibility with respect to some parametric models, since the structured variables permit the spectra to vary for each spectral band.
2) Address SV in very general variability patterns.
3) Introduce a novel manner to initialize the spectra of the proposed algorithms and the tested state-of-the-art methods.

The remainder of this paper is organized as follows. The data mixing model is described in Section II, in particular a great attention is paid to the adaptation of LMM to deal with SV by considering new variables. Section III presents the proposed algorithms derived from the introduced cost function and describes the initialization strategy adopted in the conducted experiments. Section IV presents the results obtained from the experiments conducted on synthetic and real data. Finally, Section V concludes this work.

## II. PROPOSED DATA MIXING MODEL

As stated above, within the framework of LMM, the observed nonnegative reflectance spectrum associated with a given pixel $i$ can be written as follows:

$$x_i = \sum_{j=1}^{M} r_j c_{ji} \qquad i = 1, \dots\dots, P, \tag{1}$$

where $r_j \in R_+^{L\times 1}$ denotes the nonnegative reflectance spectrum of the $j^{th}$ endmember. The coefficients $c_{ji}$ stand for the nonnegative abundance fraction of the $j^{\text{th}}$ endmember in the considered pixel $i$. $P$ and $L$ are, respectively, the number of image pixels and the number of spectral bands, while $M$ corresponds to the number of endmembers.

In the present study, SV is addressed by using an extended version of LMM introduced in [35] to model the data. Mainly, the extended model differs from LMM since it considers a different spectrum in each pixel of an HSI for each class of endmembers. For each class of endmembers, it allows one to derive different but close spectra in all the HSI pixels. Hence, this feature enables one to measure SV in spectral and spatial domains. Therefore, the model (1) becomes [35]:

$$x_i = \sum_{j=1}^{M} r_{ij} c_{ji} \qquad i = 1, \dots\dots, P. \tag{2}$$

where $r_{ij} \in R_+^{L\times 1}$ is the version associated with pixel $i$ for the $j^{\text{th}}$ class of nonnegative endmember reflectance spectra.

SV is considered by adding two vector-form parameters into $r_{ij}$ for each pixel. This permits to handle SV in the spectral part of the involved variables. According to this representation, $r_{ij}$ can be formulated as [37]-[39]:

$$r_{ij} = s_{ij} + a_{ij} - b_{ij}. \quad (3)$$

Here, $s_{ij} \in R_+^{L\times 1}$ denotes the nonnegative reflectance spectrum that is the reflectance spectrum for the $j^{\text{th}}$ class of endmembers. $a_{ij}$ and $b_{ij}$ are the nonnegative SV vectors for the version of the $j^{\text{th}}$ class of endmembers associated with pixel $i$. The vectors $a_{ij}$ and $b_{ij}$ allow the tuning of SV separately in each spectral band of the reference spectrum $s_{ij}$ within a certain interval in each pixel of the considered HSI. This feature enables one to deal with SV in a general pattern and gives high flexibility to the proposed parametric model. The model described in Eq. (2) can be reformulated by inserting the expression of $r_{ij}$ provided in Eq. (3). Therefore, Eq. (2) is formulated as follows:

$$x_i = \sum_{j=1}^{M} (s_{ij} + a_{ij} - b_{ij}) c_{ji}. \quad (4)$$

The proposed mixing model can be expressed in matrix form to describe the processed HSI by taking into consideration the particular structure of the involved variables. These structures allow one to address SV more efficiently. Thus, Eq. (4) can be written as:

$$X = (S_r K + A - B)\, C, \quad (5)$$

where $X \in R_+^{L\times P}$ is the matrix that contains the observed pixel spectra. Each row of this matrix is associated with one spectral band and each column with one observed pixel spectrum. For each pixel, to estimate a set of spectra corresponding to each class of endmembers present in the HSI, the matrix $C \in \mathbb{R}_+^{MP\times P}$ also has to be adapted. Indeed, this matrix is block-diagonal and contains the abundance fractions [35]:

$$C = \begin{bmatrix} c_1 & \begin{matrix} 0 \\ \vdots \\ 0 \end{matrix} & \cdots & \begin{matrix} 0 \\ \vdots \\ 0 \end{matrix} \\ \begin{matrix} 0 \\ \vdots \\ 0 \end{matrix} & c_2 & \ddots & \begin{matrix} 0 \\ \vdots \\ 0 \end{matrix} \\ \begin{matrix} 0 \\ \vdots \\ 0 \end{matrix} & \begin{matrix} 0 \\ \vdots \\ 0 \end{matrix} & \cdots & c_P \end{bmatrix}, \quad (6)$$

where $c_i = [c_{1i}, \dots, c_{Mi}]^{\mathrm{T}}$ is an $M$-element column vector that contains the abundance fractions associated with the $i^{\text{th}}$ observed pixel spectrum, and the notation "$(0\dots0)^{\mathrm{T}}$" corresponds to the $M$-zero-element column vector. $[.]^{\mathrm{T}}$ denotes the vector/matrix transpose. We can see clearly from the particular structure of the matrix $C$ that this enables one to extract efficiently from each pixel spectra of each class of endmembers and to tackle SV in the spatial domain. The matrix $K \in \mathbb{R}_+^{M\times MP}$ is the identity matrix of dimension $M \times M$ repeated $P$ times. The introduction of this matrix allows one both to handle SV and to avoid any additional update rule and constraint related to the particular structure of the involved matrices including $S_r$, $A$ and $B$. The matrix $S_r \in \mathbb{R}_+^{L\times M}$ contains the reference spectra. Unlike in [37]-[39], the reference spectra are not associated only with the first pixel but with the whole HSI. This undoubtedly enhances the robustness of the proposed methods, in particular when unmixing real data, which can contain, in the first pixel, an outlier and are contaminated by noise. The matrices $A \in \mathbb{R}_+^{L\times MP}$ and $B \in \mathbb{R}_+^{L\times MP}$ contain the upper and lower SV coefficients related to $a_{ij}$ and $b_{ij}$ respectively. These coefficients allow tuning the model of the whole observed spectra for each pixel of the processed HSI, with respect to the reference spectra matrix $S_r$, within an additive manner. Moreover, these coefficients delimit the upper and lower variations with respect to these considered reference spectra. Thus, $A$ and $B$ can be formulated as:

$$A = [A_1 | A_2 \dots | A_P], \quad (7)$$
with $A_t \in \mathbb{R}_+^{L\times M}$ $(t = 1,\dots,P)$,

$$B = [B_1 | B_2 \dots | B_P], \quad (8)$$
with $B_t \in \mathbb{R}_+^{L\times M}$ $(t = 1,\dots,P)$.

## III. PROPOSED UNMIXING ALGORITHMS

In this paper, two algorithms are proposed, which are pixel-by-pixel extensions of NMF. In its standard form, NMF aims at decomposing the observed HSI into the product of two nonnegative matrices. This makes NMF attractive for linear hyperspectral unmixing since it tailors well the considered data and variables [43]. Moreover, the proposed algorithms use constraints accounting for SV and the particular structures of the involved variables in order to improve the unmixing performance. The introduced algorithms are gradient descent-based ones and aim at estimating variables by modelling the data mixing function defined by (5). The involved variables are $\tilde{S_r}$, $\tilde{A}$, $\tilde{B}$ and $\tilde{C}$ that respectively represent the estimate of variables $S_r$, $A$, $B$ and $C$. The proposed approaches consist in minimizing the following cost function by using iterative and gradient-based update rules:

$$J = \frac{1}{2}\left\|X - (\tilde{S_r}K + \tilde{A} - \tilde{B})\tilde{C}\right\|_F^2 + \frac{\lambda_A}{2}\left\|\tilde{A}\right\|_F^2 + \frac{\lambda_B}{2}\left\|\tilde{B}\right\|_F^2, \quad (9)$$

where $\|.\|_F$ stands for the Frobenius norm.

The two penalization terms at the end of (9) limit the Frobenius norms of the matrices $\tilde{A}$ and $\tilde{B}$. These penalizations have two purposes. The first one is related to the general framework of NMF. Besides the limitations cited above (initialization and noise), NMF is also limited by the fact it cannot guarantee that the estimated solution is a global minimum. Therefore, the two penalization terms permit to constrain the solution space and improve the unmixing performance of the introduced algorithms. The second objective is to control the amplitude of the SV coefficients. These penalization terms allow to obtain coherent SV around the reference spectra $\tilde{S_r}$ for each pixel of the HSI. $\lambda_A$ and $\lambda_B$ are positive balancing coefficients to manage the upper and the lower SV, respectively. and do not necessarily have the same values.

Since the two algorithms are gradient descent-based methods, gradient expressions of the considered cost function are derived by making use of the properties provided in [44] and read:

$$\frac{\partial J}{\partial \widetilde{S_r}} = -X\tilde{C}^T K^T + \widetilde{S_r} K\tilde{C}\tilde{C}^T K^T + \tilde{A}\tilde{C}\tilde{C}^T K^T - \tilde{B}\tilde{C}\tilde{C}^T K^T, \quad (10)$$

$$\frac{\partial J}{\partial \tilde{A}} = -X\tilde{C}^T + \widetilde{S_r} K\tilde{C}\tilde{C}^T + \tilde{A}\tilde{C}\tilde{C}^T - \tilde{B}\tilde{C}\tilde{C}^T + \lambda_A \tilde{A}, \quad (11)$$

$$\frac{\partial J}{\partial \tilde{B}} = X\tilde{C}^T - \widetilde{S_r} K\tilde{C}\tilde{C}^T - \tilde{A}\tilde{C}\tilde{C}^T + \tilde{B}\tilde{C}\tilde{C}^T + \lambda_B \tilde{B}, \quad (12)$$

$$\frac{\partial J}{\partial \tilde{C}} = -(\widetilde{S_r}K + \tilde{A})^T X + \tilde{B}^T X - \left(\tilde{B}^T(\widetilde{S_r}K + \tilde{A}) + (\widetilde{S_r}K + \tilde{A})^T \tilde{B}\right)\tilde{C} + (\widetilde{S_r}K + \tilde{A})^T(\widetilde{S_r}K + \tilde{A})\tilde{C} + \tilde{B}^T\tilde{B}\tilde{C}. \quad (13)$$

Since each element of the matrix $\widetilde{S_r}$ has to be lower than or equal to one (i.e., when considering the relative reflectance measures), the following constraints should be considered:

$$0^{L\times M} \leq \tilde{A}_t \leq \min(u^{L\times M}, 1^{L\times M} + \tilde{B}_t - \widetilde{S_r}), \quad \forall t = 1 \ldots P, \quad (14)$$

$$0^{L\times M} \leq \tilde{B}_t \leq \min(v^{L\times M}, \tilde{A}_t + \widetilde{S_r}), \quad \forall t = 1 \ldots P, \quad (15)$$

where $u$ (respectively $v$) corresponds to the maximum value that can be reached by any element of the considered submatrix of $\tilde{A}$ (respectively $\tilde{B}$).

### A. Gradient-based SAV-LMM

The first proposed method, called Grad-SAV-LMM, applies the gradient descent approach with a fixed learning rate [42]. Thus, the following iterative update rule is applied to estimate the involved variables:

$$\theta \leftarrow \theta - \varphi_\theta \frac{\partial J}{\partial \theta}, \quad (16)$$

$\theta$ stands for one of the four estimated matrices $\widetilde{S_r}$, $\tilde{A}$, $\tilde{B}$ and $\tilde{C}$, and $\varphi_\theta$ is a fixed scalar learning rate. According to (16), the update rules, for the four considered variables, are given by:

$$\widetilde{S_r} \leftarrow \widetilde{S_r} - \varphi_{\widetilde{S_r}} \frac{\partial J}{\partial \widetilde{S_r}}, \quad (17)$$

$$\tilde{A} \leftarrow \tilde{A} - \varphi_{\tilde{A}} \frac{\partial J}{\partial \tilde{A}}, \quad (18)$$

$$\tilde{B} \leftarrow \tilde{B} - \varphi_{\tilde{B}} \frac{\partial J}{\partial \tilde{B}}, \quad (19)$$

$$\tilde{C} \leftarrow \tilde{C} - \varphi_{\tilde{C}} \frac{\partial J}{\partial \tilde{C}}. \quad (20)$$

The convergence criterion is defined by:

$$\left|\frac{J^{(t)} - J^{(t+1)}}{J^{(t)}}\right| \leq Tresh, \quad (21)$$

where $t$ is an iteration and $Tresh$ is a predefined threshold. Moreover, to avoid an excessive number of iterations, a maximum number of iterations is also considered.

### B. Multiplicative-based SAV-LMM

The second proposed algorithm is based on the multiplicative gradient based-updated rules, which are obtained by the following procedure. The gradient expressions of $J$ with respect to a matrix variable $\theta$ can here be formulated as the difference of two nonnegative functions [4], such that $\frac{\partial J}{\partial \theta} = \frac{\partial J}{\partial \theta}^+ - \frac{\partial J}{\partial \theta}^-$. The functions $\frac{\partial J}{\partial \theta}^+$ and $\frac{\partial J}{\partial \theta}^-$ correspond to the terms of (10)-(13) preceded by a plus and minus sign, respectively. The nonnegativity of $\theta$ throughout its adaptation can be fulfilled by initializing $\theta$ with a nonnegative value and by properly choosing the value of the learning rate matrix $\varphi_\theta$ according to:

$$\varphi_\theta = \theta \oslash \frac{\partial J}{\partial \theta}^+, \quad (22)$$

where $\oslash$ stands for the element-wise division. Therefore, the update rule (16) is here modified by using the element-wise multiplication by the matrix $\varphi_\theta$ and thus becomes [4]

$$\theta \leftarrow \theta \odot \frac{\partial J}{\partial \theta}^- \oslash \frac{\partial J}{\partial \theta}^+. \quad (23)$$

Thus, the proposed multiplicative update rules, for the four considered variables, are:

$$\widetilde{S_r} \leftarrow \widetilde{S_r} \odot (X\tilde{C}^T K^T + \tilde{B}\tilde{C}\tilde{C}^T K^T) \oslash (\widetilde{S_r} K\tilde{C}\tilde{C}^T K^T + \tilde{A}\tilde{C}\tilde{C}^T K^T + \varepsilon), \quad (24)$$

$$\tilde{A} \leftarrow \tilde{A} \odot (X\tilde{C}^T + \tilde{B}\tilde{C}\tilde{C}^T) \oslash (\widetilde{S_r} K\tilde{C}\tilde{C}^T + \tilde{A}\tilde{C}\tilde{C}^T + \lambda_A \tilde{A} + \varepsilon), \quad (25)$$

$$\tilde{B} \leftarrow \tilde{B} \odot (\widetilde{S_r} K\tilde{C}\tilde{C}^T + \tilde{A}\tilde{C}\tilde{C}^T) \oslash (X\tilde{C}^T + \tilde{B}\tilde{C}\tilde{C}^T + \lambda_B \tilde{B} + \varepsilon), \quad (26)$$

$$\tilde{C} \leftarrow \tilde{C} \odot ((\widetilde{S_r}K + \tilde{A})^T X + \left(\tilde{B}^T(\widetilde{S_r}K + \tilde{A}) + (\widetilde{S_r}K + \tilde{A})^T\tilde{B}\right)\tilde{C}) \oslash (\tilde{B}^T X + (\widetilde{S_r}K + \tilde{A})^T(\widetilde{S_r}K + \tilde{A})\tilde{C} + \tilde{B}^T\tilde{B}\tilde{C} + \varepsilon), \quad (27)$$

where $\varepsilon$ is a very small and positive value added to each denominator of update rules (24)-(27) to prevent any division by zero.

### C. Initialization by AEEB

As mentioned above, the performance of the proposed algorithms (and also of the considered state-of-art-techniques), which are nonconvex optimization problems, depend on their initialization. To improve this aspect, a straightforward strategy is adopted by using a (pseudo or image-based) spectral library.

To this end, such a spectral library is constructed by means of AEEB [18]. AEEB relies on a simple assumption that statistics of the considered HSI can be approximately recovered by using some small parts of it. Indeed, if there are a sufficient number of pure pixels (pixels containing only one endmember), they can be used to describe SV present in the observed scene by providing suitable candidates for each class of endmembers present in the observed HSI. One of the main motivations behind the use of such a spectral library is the fact that it can handle different types of SV [6]. The initialization proposed here is divided in three main stages. The first two stages construct the spectral library like in [18]. The last stage is devoted to the selection of the most suitable spectra to represent a class of endmembers.

More precisely, the first stage consists in building spectra bundles by means of several runs of an EEA on several randomly selected subsets of the HSI. In our experiments, the Vertex Component Analysis (VCA) algorithm [45] is used as an EEA. In the second stage, the extracted spectra are clustered into their corresponding class of endmembers. Such a clustering is achieved by using the *k-means* algorithm that considers the spectral angle distance or cosine distance [46]. The use of this distance is motivated by the fact that it is scale invariant [47]. The third and last stage consists in assigning the representative spectrum of each class endmember. For the best candidate for each class of pure materials, two strategies are considered:

1) Init-a: the candidate of each class is obtained by averaging the spectra composing each class of endmembers.

2) Init-b: the candidate of the associated class of endmembers is derived by selecting the median spectrum of each class of endmembers.

Moreover, the use of AEEB implies to set some parameters *a priori* including the number of pure materials present in each subset, the number of selected subsets and their corresponding size. For the sake of clarity, the performance of AEEB is directly related to the presence of a sufficient number of pure pixels in the processed HSI.

The Grad-SAV-LMM and Multi-SAV-LMM algorithms are described in Algorithm 1 and Algorithm 2, respectively.

**Algorithm 1:** Grad-SAV-LMM.

**Input:** Hyperspectral image $X$

***1. Initialization stage***

1.1- Initialize $\tilde{S_r}$ from $X$ by using the VCA method or by Init-a or Init-b.

1.2- Set each element of $\tilde{C}$ by means of FCLS.

**2.** ***Optimization stage*** (until reaching the predefined maximum number of iterations or satisfying (21))

2.1- Optimize $\tilde{S_r}$ by using (17).

2.2- Optimize $\tilde{A}$ by using (18) and (14).

2.3- Optimize $\tilde{B}$ by using (19) and (15).

2.4- Optimize $\tilde{C}$ by using (20).

**Output:** The estimated spectra of each class of endmembers and their associated abundance coefficients

**Algorithm 2:** Multi-SAV-LMM.

**Input:** Hyperspectral image $X$

***1. Initialization stage***

1.1- Initialize $\tilde{S_r}$ from $X$ by using the VCA method or by Init-a or Init-b.

1.2- Set each element of $\tilde{C}$ by means of FCLS.

**2.** ***Optimization stage*** (until reaching the predefined maximum number of iterations or satisfying (21))

2.1- Optimize $\tilde{S_r}$ by using (24).

2.2- Optimize $\tilde{A}$ by using (25) and (14).

2.3- Optimize $\tilde{B}$ by using (26) and (15).

2.4- Optimize $\tilde{C}$ by using (27).

**Output:** The estimated spectra of each class of endmembers and their associated abundance coefficients.

## IV. TEST RESULTS

### *A. Experimental setup and performance criteria*

In this section, the performances of the designed algorithms are compared to some state-of-art methods. For the conducted experiments, we consider four methods, which consider SV, where their implementation details are briefly described in the following.

1) IP-NMF [35]: the endmembers are initialized by VCA, Init-a and Init-b. As suggested by the authors in [35], the abundances are initialized by FCLS [48] and we fix the regularization parameter $\mu$ to 30.
2) PLMM [26]: the abundances are initialized by FCLS, we set $\gamma = 1$ as suggested by the authors in [26]. $\alpha$ and $\beta$ are respectively set to 0.001 and 500. The endmember spectra are also initialized by VCA, Init -a and Init -b.
3) ELMM [27]: the endmember spectra are also initialized by VCA, Init-a and Init-b. The abundances are initialized by FCLS. We fix $\lambda_S$, $\lambda_A$ and $\lambda_\psi$ to, respectively, 0.1, 0.01 and 0.001.
4) GLMM [28]: we set $\lambda_M$, $\lambda_A$ and $\lambda_\psi$ to 0.1, 0.01 and 0.001 respectively. The endmember spectra are also initialized by VCA, Init-a and Init-b. The abundances are initialized by FCLS.

For the proposed techniques, we set the regularization parameters $\lambda_A$ and $\lambda_B$ in the following interval [0.1, 0.2, 0.4, 1, 2, 3, 4, 5, 10].

To assess the performances on the semi-synthetic dataset, the following measures are considered: Spectral Angle Mapper ($SAM_S$), Spectral Information Divergence ($SID_S$), and Normalized Mean Square Error ($NMSE_S$), calculated between reference and estimated spectra, defined respectively by:

$$SAM_S = \frac{1}{PM}\sum_{i=1}^{P}\sum_{j=1}^{M} arccos\left(\frac{\langle r_{ij}, \tilde{r}_{ij}\rangle}{\|r_{ij}\|_2 \cdot \|\tilde{r}_{ij}\|_2}\right), \quad (28)$$

$$SID_S = \frac{1}{PM}\sum_{i=1}^{P}\sum_{j=1}^{M}\left|r_{ij}^T ln\,(r_{ij} \oslash \tilde{r}_{ij}) + \tilde{r}_{ij}^T ln\,(\tilde{r}_{ij} \oslash r_{ij})\right|, \quad (29)$$

$$NMSE_S = \frac{1}{PM}\sum_{i=1}^{P}\sum_{j=1}^{M}\frac{\|r_{ij}-\tilde{r}_{ij}\|_2}{\|r_{ij}\|_2}, \quad (30)$$

We also consider the abundance overall Root Mean Square Error ($RMSE_A$), calculated between reference and estimated linear abundance fraction maps, defined by:

$$RMSE_A = \frac{1}{P}\sum_{i=1}^{P}\sqrt{\frac{1}{M}\sum_{j=1}^{M}(c_{ji}-\tilde{c}_{ji})^2}, \qquad (31)$$

For the real data, since there is no ground truth, the following performance criteria are used: The reconstruction Spectral Angle Mapper ($SAM_X$), and the reconstruction RMSE ($RMSE_X$), calculated between the considered original and reconstructed HSI images.

$$SAM_X = \frac{1}{P}\sum_{i=1}^{P} arcos\left(\frac{\langle x_{.i}, \tilde{x}_{.i}\rangle}{\|x_{.i}\|\|\tilde{x}_{.i}\|}\right), \qquad (32)$$

$$RMSE_X = \frac{1}{P}\sum_{i=1}^{P}\sqrt{\frac{1}{L}\sum_{l=1}^{L}(x_{li}-\tilde{x}_{li})^2}. \qquad (33)$$

For all the used criteria, the lower their values, the better the tested method.

## *B. Test results with semi-synthetic data*

For the conducted experiments, two spectral libraries are used. The first spectral library (SL-1) (see Fig. 1(a)) is composed of three classes of urban environment spectra obtained from ground measurements using a spectrometer during the French airborne UMBRA (Urban Material characterization in the sun and shade of Built-up structures and trees and their Retrieval from Airborne image) campaign [49]. This spectral library highlights well SV that can be present in real conditions particularly in urban scene. The three classes correspond to photovoltaic panels, tiles and grass. The pure material spectra contain 214 spectral bands after removing the atmospheric absorption bands [50] ranging from 400 to 2500 nm.

Five pure materials compose the second spectral library (SL-2) (see Fig. 1(b)), namely: railway, water, running track, healthy and synthetic grass. The considered hyperspectral endmember spectra with SV are derived from ground truth samples provided in [52]. These spectra, with 144 wavelengths, cover the spectral domain 0.36-1.04 µm.

These two spectral libraries enable one to generate synthetic data which are fairly representative of SV present in real HSIs. The first synthetic hyperspectral image (HI-1), generated using SL-1, contains 30×30 pixels. The abundances are obtained with the HYDRA toolbox [51].

The second synthetic hyperspectral image (HI-2) is generated following the same procedure as with HI-1, now using SL-2.

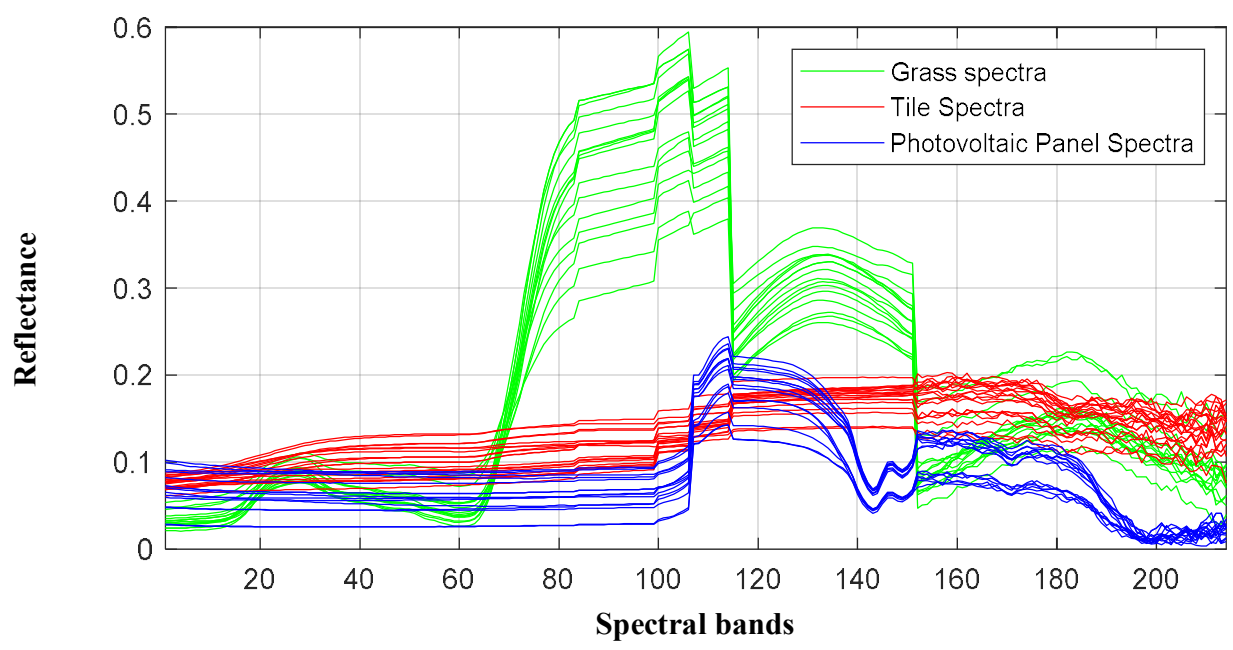

**(a)**

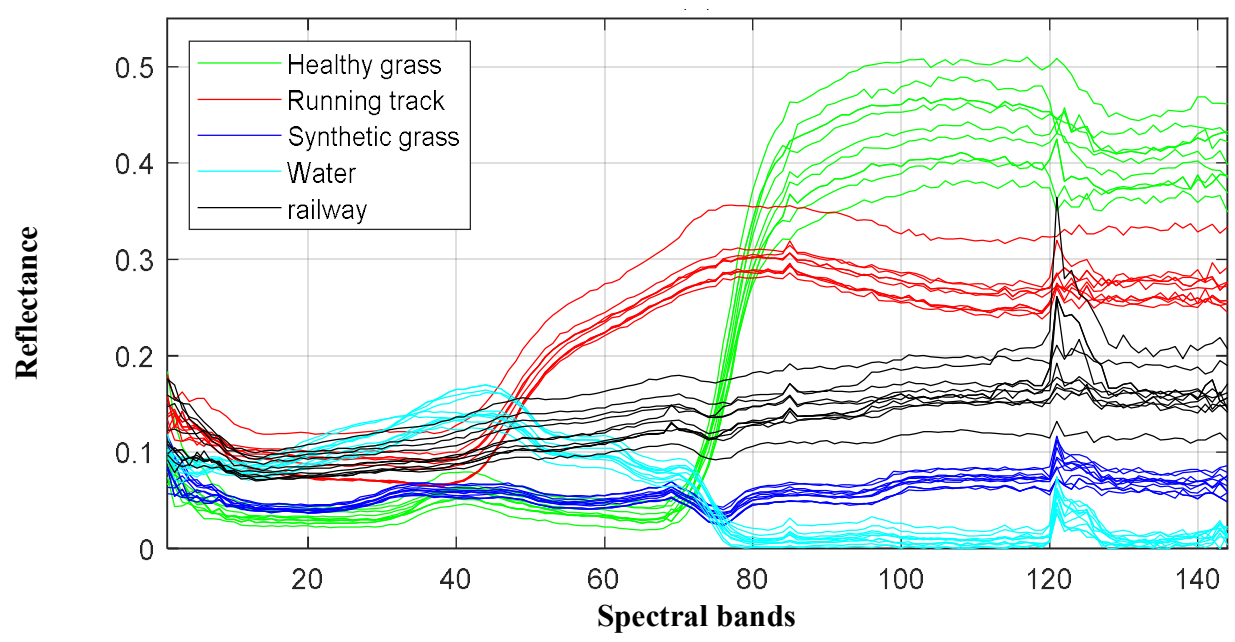

**(b)**

**Fig. 1** Used spectral libraries in semi-synthetic data experiments: (a) SL-1, and (b) SL-2.

The considered semi-synthetic hyperspectral images are corrupted by white Gaussian noise with different signal-to-noise ratios (SNR) including 20, 25 and 30 dB.

The results for each configuration of noise level and used initialization for HI-1 are reported in Table I. The best results are shown in bold red and the second ones are in bold blue. First of all, for the classical initialization with VCA, we can see clearly that ELMM yields highest values of $SAM_S$ in the configuration with a 20-dB SNR which demonstrates that SV is not only induced by topography and illumination changes, in particular in urban scenes like in our scenario. GLMM performs better since it has more flexibility. IP-NMF reaches even better values: this proves the effectiveness of the used extended version of LMM which considers SV in both spatial and spectral parts.

Table I demonstrates that the proposed initialization strategies (Init-a and Init-b) improve the performance of all tested methods, with substantial gains in both the spectral and spatial domains compared with the standard VCA-based initialization, particularly for HI-1 at SNR = 20 dB. These results confirm that considering SV during the initialization process reduces the impact of noise and decreases the sensitivity of the methods to their initialization.

Moreover, Table I shows that both proposed Grad-SAV-LMM and Multi-SAV-LMM methods outperform the state-of-the-art techniques in almost all tested configurations. Grad-SAV-LMM and Multi-SAV-LMM achieve the best results according to the considered spectral criteria, reaching the lowest values of $SAM_S$. Their significant performance improvements further highlight their robustness. In addition, Multi-SAV-LMM provides the best spatial performance, with the lowest values of $RMSE_A$ across almost all SNR levels.

Indeed, the results in Table I show that the Grad-SAV-LMM and Multi-SAV-LMM methods generally offer the best performance for HI-1 hyperspectral unmixing. In particular, Grad-SAV-LMM generally achieves the lowest mean values for the considered evaluation criteria (example for Init-b and SNR=20 dB: $SAM_S = 5.61°$, $SID_S = 1.00$, $NMSE_S = 15.25$ % and $RMSE_A = 0.0943$), while Multi-SAV-LMM often yields the best mean values in terms of the spatial metrics considered (example for Init-b and SNR=20: $SAM_S = 6.11°$, $SID_S = 1.15$, $NMSE_S = 16.23$ % and $RMSE_a = 0.0896$). The improvement is particularly noticeable when the SNR increases from 20 to 30 dB, confirming the robustness of both approaches with respect to noise reduction.

Fig. 2 illustrates the obtained abundance maps for each method in the case of SNR=30 dB and VCA initialization. We can see that Multi-SAV-LMM retrieves better the abundances in particular for the first and third endmembers.

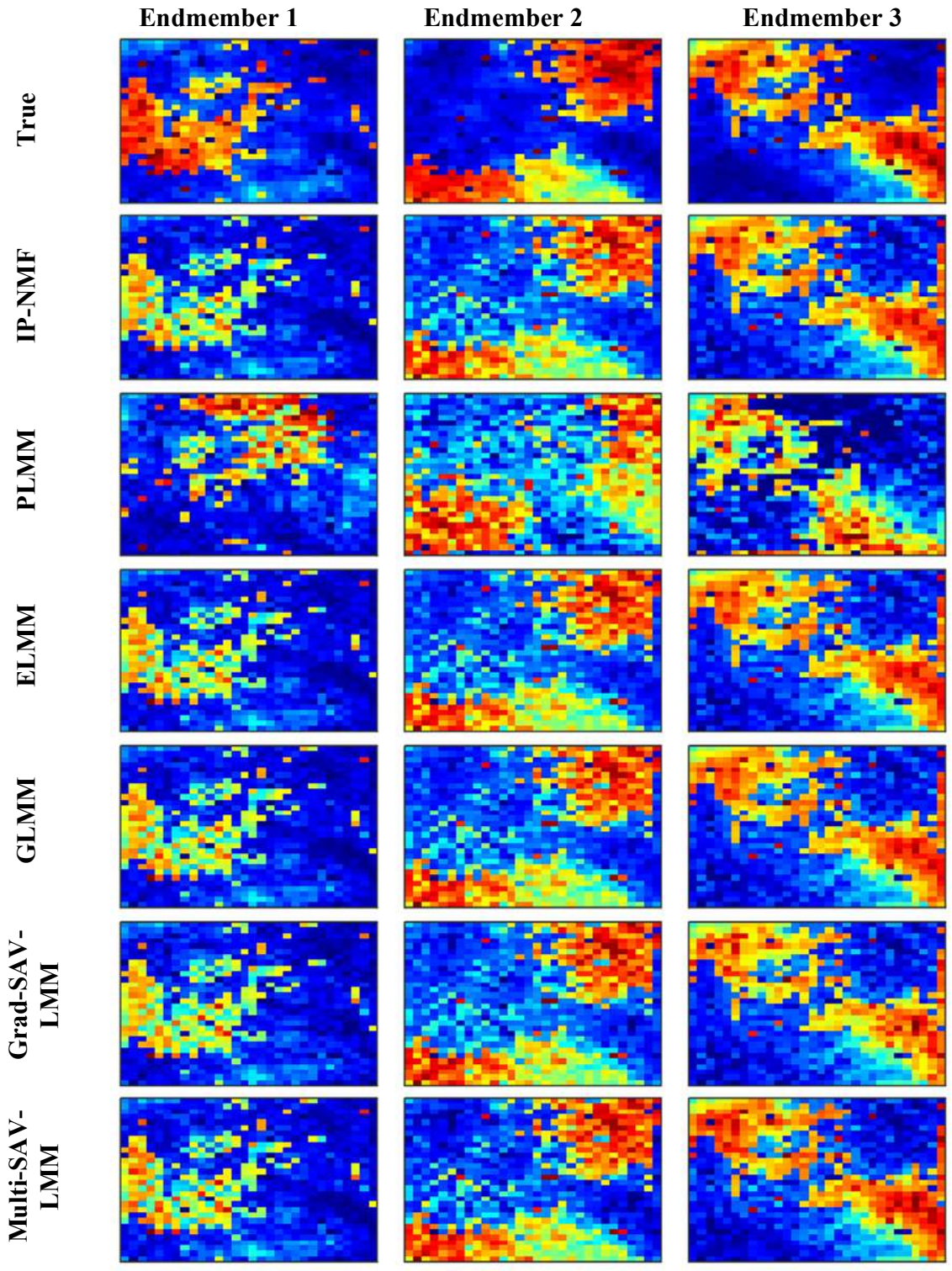

**Fig. 2** Obtained abundances maps for HI-1 using different techniques for SNR=30dB.

Table II reports the results obtained with the HI-2 dataset. Similar to what was observed for HI-1, the Init-a and Init-b strategies improve the performances of all considered state-of-the-art methods, yielding better results in both spectral and spatial domains compared with the standard VCA-based initialization. These findings confirm that considering SV during the initialization process enhances the robustness of the tested approaches, particularly at high SNR levels.

Overall, the results reported in Table II show that the proposed Multi-SAV-LMM and Grad-SAV-LMM methods achieve the most competitive performance across all considered spectral and spatial criteria. For example, in the configuration with Init-b and $SNR = 20dB$, the Grad-SAV-LMM method achieves the following performance: $SAM_S = 4.25°$, $SID_S = 0.58$, $NMSE_S = 12.21\ \%$ and $RMSE_a = 0.1001$, the Multi-SAV-LMM method achieves the following performance: $SAM_S = 4.31°$, $SID_S = 0.60$, $NMSE_S = 12.26\ \%$ and $RMSE_a = 0.0999$. Besides, the IP-NMF and PLMM methods remain competitive in certain configurations. The consistent improvements achieved with Init-a and Init-b once again demonstrate the effectiveness of the proposed initialization strategies.

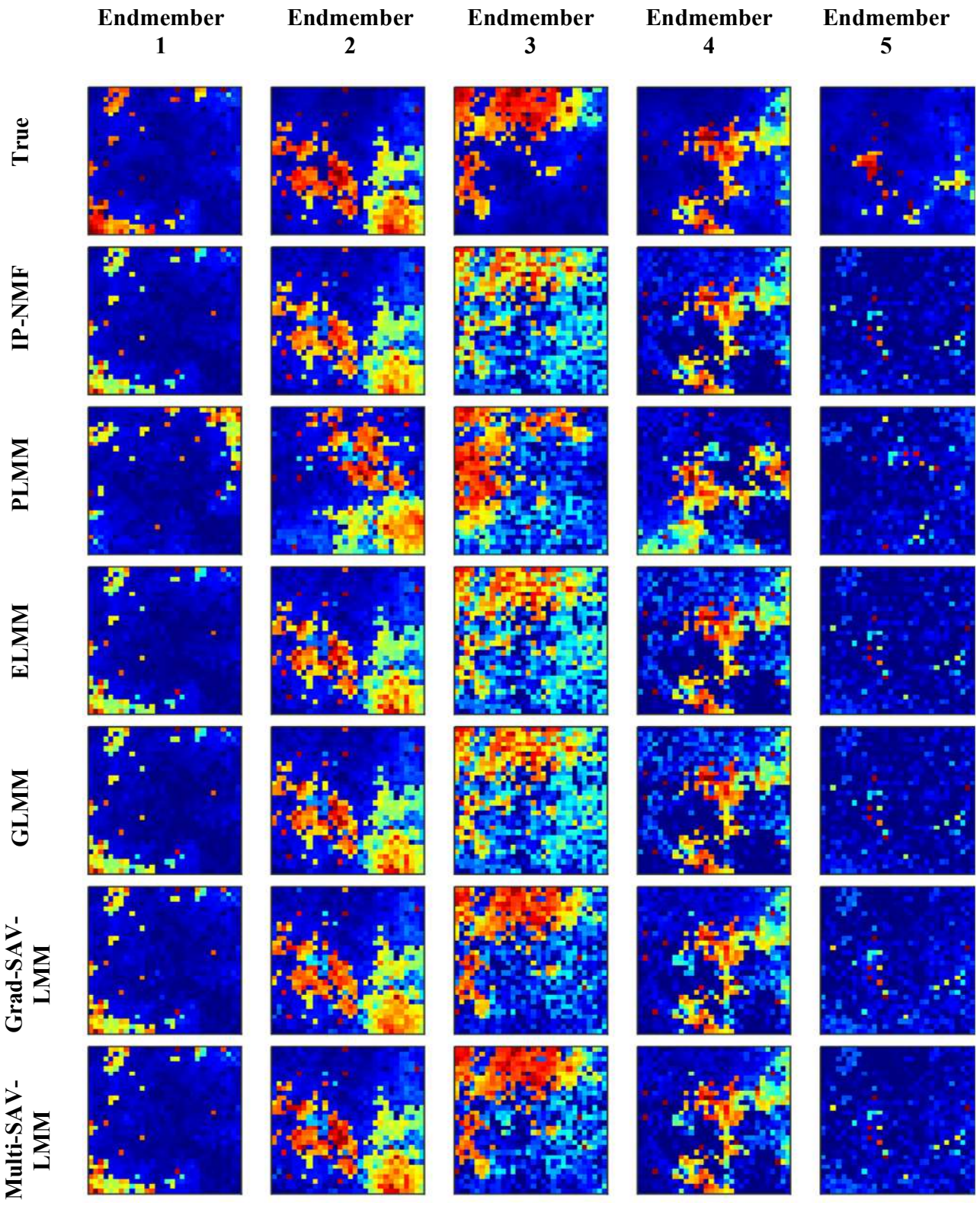

**Fig. 3** Obtained abundances maps for HI-2 using different techniques for SNR=30dB.

Fig. 3 shows the estimated abundance maps for each method initialized by VCA in the case of SNR = 30 dB. As in the case of HI-1, PLMM fails to accurately retrieve the abundance maps. ELMM and GLMM provide more accurate estimations than PLMM. However, Fig. 3 illustrates that the proposed approaches produce superior results, particularly for Endmember 3, where the abundance maps generated by the proposed methods are closer to the ground truth compared to those produced by the other state-of-the-art techniques.

*C. Test results with real data*

Two real hyperspectral images are considered in the conducted experiments. The first one (RHI-1) was acquired over the University of Houston, more precisely the Robertson stadium on the Houston Campus [30]. RHI-1 has 144 spectral bands in the 380 nm to 1050 nm spectral domain and is composed of 105×128 pixels. RHI-1 is constituted of four main classes of endmembers which are vegetation, concrete stands, asphalt and red metallic roofs [30]. The second considered real hyperspectral image (RHI-2) is the Jasper Ridge dataset. More precisely, it is a subset composed of 70×70 pixels and has 224 bands covering the wavelengths from 380 nm to 2500 nm. After removing channels 1–3, 108–112, 154–166, and 220–224 due to strong water vapor absorption and atmospheric effects, 198 spectral bands remain. This image is constituted of four main pure classes of materials, namely: water, dirt, tree and road.

**TABLE I**

MEANS OF THE PERFORMANCE CRITERIA FOR THE HI-1 IMAGE

| Data | SNR | Initialization | Criterion | IP-NMF | PLMM | ELMM | GLMM | Grad-SAV-LMM | Multi-SAV-LMM |
|---|---|---|---|---|---|---|---|---|---|
| HI-1 | 20 dB | VCA | $SAM_S$ (°) | 7.55 | 7.91 | 8.67 | 7.93 | **6.54** | **6.60** |
| | | | $SID_S$ | 3.73 | 2.10 | 2.56 | **1.68** | 2.80 | **1.40** |
| | | | $NMSE_S$ (%) | 23.58 | 22.94 | 20.69 | 18.09 | **17.36** | **17.45** |
| | | | $RMSE_A$ | 0.116 | 0.3699 | **0.1027** | 0.1035 | 0.1133 | **0.0992** |
| | | Init-a | $SAM_S$ (°) | **5.83** | 6.32 | 7.34 | 7.45 | **5.63** | 6.17 |
| | | | $SID_S$ | 1.39 | 1.24 | 1.56 | 1.50 | **1.01** | **1.18** |
| | | | $NMSE_S$ (%) | 18.28 | 17.68 | 17.78 | 17.37 | **15.29** | **16.37** |
| | | | $RMSE_A$ | 0.0990 | 0.3863 | **0.0975** | **0.0975** | 0.0977 | **0.0945** |
| | | Init-b | $SAM_S$ (°) | **5.86** | 6.32 | 7.35 | 7.42 | **5.61** | 6.11 |
| | | | $SID_S$ | 1.36 | 1.23 | 1.58 | 1.48 | **1.00** | **1.15** |
| | | | $NMSE_S$ (%) | 17.95 | 17.42 | 17.74 | 17.28 | **15.25** | **16.23** |
| | | | $RMSE_A$ | 0.0949 | 0.3848 | 0.0929 | **0.0925** | 0.0943 | **0.0896** |
| | 25 dB | VCA | $SAM_S$ (°) | 7.06 | 7.22 | 7.43 | 7.14 | **6.53** | **6.56** |
| | | | $SID_S$ | 2.16 | 1.90 | 2.24 | **1.74** | 1.90 | **1.68** |
| | | | $NMSE_S$ (%) | 18.59 | 19.22 | 17.68 | **17.14** | **17.39** | 17.79 |
| | | | $RMSE_A$ | 0.1084 | 0.3825 | 0.1062 | **0.1048** | 0.1050 | **0.0978** |
| | | Init-a | $SAM_S$ (°) | 5.69 | 5.95 | 6.08 | 6.13 | **5.43** | **5.61** |
| | | | $SID_S$ | 1.36 | 1.23 | 1.33 | 1.18 | **0.98** | **1.11** |
| | | | $NMSE_S$ (%) | 16.33 | 16.44 | 15.93 | 15.40 | **14.23** | **15.32** |
| | | | $RMSE_A$ | 0.0888 | 0.3991 | 0.0879 | 0.0876 | **0.0871** | **0.0850** |
| | | Init-b | $SAM_S$ (°) | 5.79 | 5.98 | 6.17 | 6.26 | **5.49** | **5.78** |
| | | | $SID_S$ | 1.55 | 1.33 | 1.43 | 1.25 | **1.02** | **1.13** |
| | | | $NMSE_S$ (%) | 17.28 | 16.60 | 16.42 | 15.84 | **14.31** | **15.71** |
| | | | $RMSE_A$ | 0.0922 | 0.4039 | 0.0908 | 0.0903 | **0.0892** | **0.0873** |
| | 30 dB | VCA | $SAM_S$ (°) | 7.00 | 7.23 | 6.98 | 6.64 | **6.32** | **6.40** |
| | | | $SID_S$ | 2.44 | 2.08 | 2.31 | **1.63** | 1.87 | **1.74** |
| | | | $NMSE_S$ (%) | 19.46 | 19.44 | 17.52 | **16.62** | **17.21** | 17.78 |
| | | | $RMSE_A$ | 0.1070 | 0.3706 | 0.1054 | **0.1013** | 0.1042 | **0.0989** |
| | | Init-a | $SAM_S$ (°) | 5.67 | 6.38 | 5.48 | 5.56 | **5.27** | **5.43** |
| | | | $SID_S$ | 1.46 | 1.33 | 1.26 | **1.09** | **0.98** | **1.09** |
| | | | $NMSE_S$ (%) | 16.63 | 18.51 | 15.14 | **14.61** | **13.98** | 15.57 |
| | | | $RMSE_A$ | 0.0885 | 0.3843 | 0.0875 | 0.0871 | **0.0866** | **0.0855** |
| | | Init-b | $SAM_S$ (°) | 5.72 | 5.87 | **5.51** | 5.59 | **5.29** | 5.52 |
| | | | $SID_S$ | 1.63 | 1.39 | 1.35 | **1.10** | **0.99** | **1.10** |
| | | | $NMSE_S$ (%) | 17.71 | 17.04 | 15.75 | **14.86** | **14.12** | 15.86 |
| | | | $RMSE_A$ | 0.0889 | 0.4074 | 0.0879 | 0.0872 | **0.0863** | **0.0850** |

**TABLE II**
MEANS OF THE PERFORMANCE CRITERIA FOR THE HI-2 IMAGE

| Data | SNR | Initialization | Criterion | IP-NMF | PLMM | ELMM | GLMM | Grad-SAV-LMM | Multi-SAV-LMM |
|---|---|---|---|---|---|---|---|---|---|
| HI-2 | 20 dB | VCA | $SAM_S$ (°) | **4.90** | 5.43 | 6.85 | 7.81 | **4.72** | 5.34 |
| | | | $SID_S$ | 1.19 | 1.05 | 1.15 | 1.37 | **0.98** | **0.83** |
| | | | $NMSE_S$ (%) | 18.07 | 17.78 | 17.83 | 18.74 | **17.45** | **14.21** |
| | | | $RMSE_A$ | **0.1104** | 0.3249 | 0.1182 | 0.1170 | 0.1209 | **0.107** |
| | | Init-a | $SAM_S$ (°) | **4.45** | 4.88 | 6.42 | 7.21 | **4.20** | 4.80 |
| | | | $SID_S$ | 0.76 | 0.69 | 1.04 | 1.26 | **0.62** | **0.66** |
| | | | $NMSE_S$ (%) | 13.29 | 13.13 | 17.39 | 17.74 | **12.52** | **12.09** |
| | | | $RMSE_A$ | 0.1045 | 0.3486 | 0.1186 | 0.1176 | **0.1013** | **0.0946** |
| | | Init-b | $SAM_S$ (°) | 4.73 | 5.09 | 6.59 | 7.26 | **4.25** | **4.31** |
| | | | $SID_S$ | 0.77 | 0.65 | 1.00 | 1.23 | **0.58** | **0.60** |
| | | | $NMSE_S$ (%) | 13.16 | 12.94 | 16.72 | 17.13 | **12.21** | **12.26** |
| | | | $RMSE_A$ | 0.101 | 0.3480 | 0.1171 | 0.1162 | **0.1001** | **0.0999** |
| | 25 dB | VCA | $SAM_S$ (°) | 4.99 | 4.97 | 6.12 | 6.64 | **4.35** | **4.81** |
| | | | $SID_S$ | 0.92 | 0.65 | 0.91 | 1.10 | **0.57** | **0.71** |
| | | | $NMSE_S$ (%) | 14.43 | 13.04 | 16.31 | 18.30 | **12.19** | **12.91** |
| | | | $RMSE_A$ | 0.1152 | 0.3261 | 0.1160 | 0.1149 | **0.0902** | **0.0942** |
| | | Init-a | $SAM_S$ (°) | **4.25** | 4.37 | 5.25 | 5.47 | **4.25** | **4.37** |
| | | | $SID_S$ | 0.68 | 0.57 | 0.73 | 0.73 | **0.49** | **0.55** |
| | | | $NMSE_S$ (%) | 13.23 | 12.33 | 14.08 | 13.58 | **11.71** | **11.93** |
| | | | $RMSE_A$ | 0.1007 | 0.3400 | 0.1035 | 0.1031 | **0.0897** | **0.0897** |
| | | Init-b | $SAM_S$ (°) | **4.27** | 4.34 | 5.31 | 5.90 | **4.16** | 4.28 |
| | | | $SID_S$ | 0.6765 | 0.5581 | 0.73 | 0.75 | **0.46** | **0.51** |
| | | | $NMSE_S$ (%) | 13.58 | 12.35 | 14.15 | 14.19 | **11.60** | **11.68** |
| | | | $RMSE_A$ | 0.1055 | 0.3404 | 0.1073 | 0.1068 | **0.0887** | **0.0893** |
| | 30 dB | VCA | $SAM_S$ (°) | 5.09 | 5.06 | 5.44 | 5.35 | **4.66** | **4.52** |
| | | | $SID_S$ | 1.06 | 0.76 | 1.07 | 0.83 | **0.96** | **0.77** |
| | | | $NMSE_S$ (%) | 15.30 | 13.10 | 16.06 | 15.64 | **12.878** | **13.00** |
| | | | $RMSE_A$ | 0.1149 | 0.32 | 0.1194 | 0.1178 | **0.0923** | **0.0932** |
| | | Init-a | $SAM_S$ (°) | **4.15** | 4.21 | 4.60 | 4.79 | 4.23 | **4.06** |
| | | | $SID_S$ | 0.68 | 0.55 | 0.66 | 0.63 | **0.50** | **0.51** |
| | | | $NMSE_S$ (%) | 13.76 | 12.65 | 13.47 | 12.80 | **12.17** | **11.31** |
| | | | $RMSE_A$ | 0.09723 | 0.3338 | 0.09946 | 0.09906 | **0.0850** | **0.0832** |
| | | Init-b | $SAM_S$ (°) | **4.31** | 4.34 | 4.75 | 4.92 | 4.35 | **4.10** |
| | | | $SID_S$ | 0.78 | 0.61 | 0.70 | 0.67 | **0.55** | **0.58** |
| | | | $NMSE_S$ (%) | 14.77 | 13.42 | 13.71 | 13.45 | **12.78** | **11.89** |
| | | | $RMSE_A$ | 0.1006 | 0.3354 | 0.1023 | 0.1018 | **0.0841** | **0.0834** |

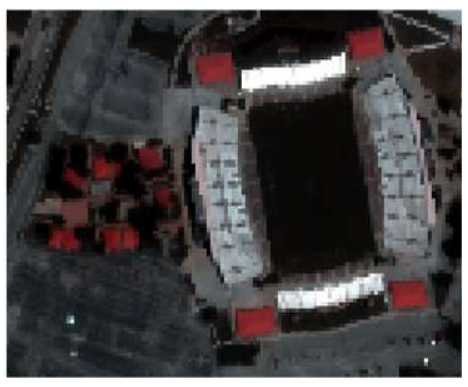

RGB representation RHI-1

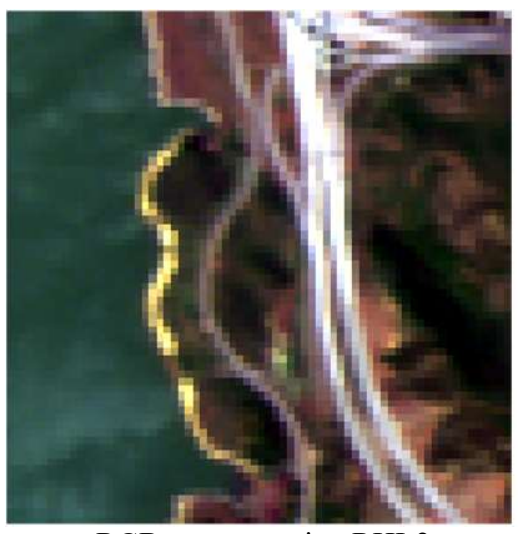

RGB representation RHI-2

**Fig. 4** True color composite representation of the used real HSIs.

Moreover, since Init-a proves to be highly effective in terms of performances, it is used as the initialization for all the tested methods for the real data. Table III presents the results obtained by the tested methods on both real datasets.

The results show that GLMM consistently achieves the best performance for both the RHI-1 and RHI-2 real datasets. For RHI-1, GLMM achieves the lowest average values, with a $SAM_X$ equal to 0.10° and a $RMSE_X$ equal to 0.0005. It is followed by ELMM, with a $SAM_X$ equal to 0.21° and a $RMSE_X$ equal to 0.0007.

For RHI-2, the same trend is observed: GLMM continues to perform best with a $SAM_X$ equal to 0.29° and a $RMSE_X$ equal to 0.0006, while ELMM ranks second a $SAM_X$ equal to 0.51° and a $RMSE_X$ equal to 0.0019. Conversely, IP-NMF yields the highest values for both datasets, indicating the poorest performance among the compared methods.

However, these considered criteria are not entirely representative of the overall performances. Indeed, it should be noted that reconstruction criteria are not always reliable indicators, as it is possible to inaccurately estimate the spectra and abundances, yet still obtain a good product of the two matrices.

Regarding the proposed methods, the following performance was obtained: for the RHI-1 real dataset, the Grad-SAV-LMM method achieved an average value of $SAM_X$ equal to 1.70° and an average value of $RMSE_X$ equal to 0.0131, while the Multi-SAV-LMM method achieved a mean $SAM_X$ equal to 1.48° and a mean $RMSE_X$ equal to 0.0048. The Multi-SAV-LMM method thus performs better than the Grad-SAV-LMM for that image, particularly for the $RMSE_X$ criterion. For RHI-2, the values increase for both methods: an average value of $SAM_X$ equal to 3.79° and a $RMSE_X$ equal to 0.0223 for Grad-SAV-LMM, compared to a $SAM_X$ equal to 4.17° and a $RMSE_X$ equal to 0.0240 for Multi-SAV-LMM.

TABLE III

MEANS OF THE PERFORMANCE CRITERIA FOR RHI-1 AND RHI-2 IMAGES

| Data | Initialization | Criterion | IP-NMF | PLMM | ELMM | GLMM | Grad-SAV-LMM | Multi-SAV-LMM |
|---|---|---|---|---|---|---|---|---|
| RHI-1 | Init-a | $SAM_X$ (°) | 2.78 | 0.82 | **0.21** | **0.10** | 1.70 | 1.48 |
| | | $RMSE_X$ | 0.0163 | 0.0037 | **0.0007** | **0.0005** | 0.0131 | 0.0048 |
| RHI-2 | Init-a | $SAM_X$ (°) | 5.65 | 2.23 | **0.51** | **0.29** | 3.79 | 4.17 |
| | | $RMSE_X$ | 0.0282 | 0.0110 | **0.0019** | **0.0006** | 0.0223 | 0.0240 |

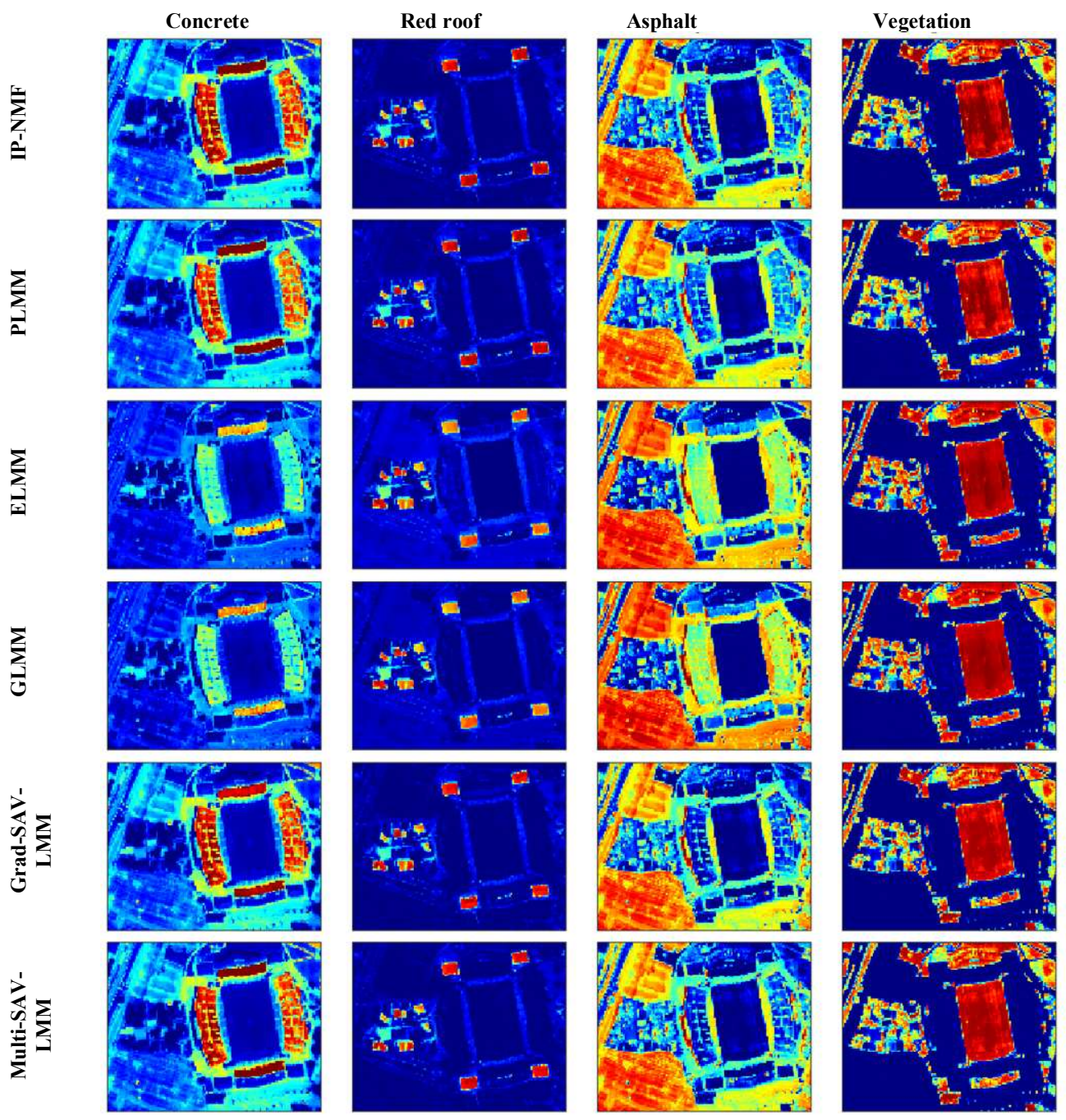

**Fig. 5** Obtained abundance maps for RHI-1.

The abundance maps estimated by each method are shown in Fig. 5 and Fig. 6. For RHI-1, the estimated abundance maps demonstrate the effectiveness of the proposed approaches. In particular, the superiority of the introduced techniques is evident in the representation of concrete and asphalt, where both proposed methods produce results that closely resemble the RGB representation of RHI-1 provided in Fig. 4.

Indeed, asphalt and concrete are more difficult to distinguish when using GLMM and ELMM, on these specific processed real-world data, highlighting the advantage of the proposed approaches in capturing spectral differences. This confirms that the proposed model and associated methods are more able to handle SV and better adapted to the complexity of real urban scenes.

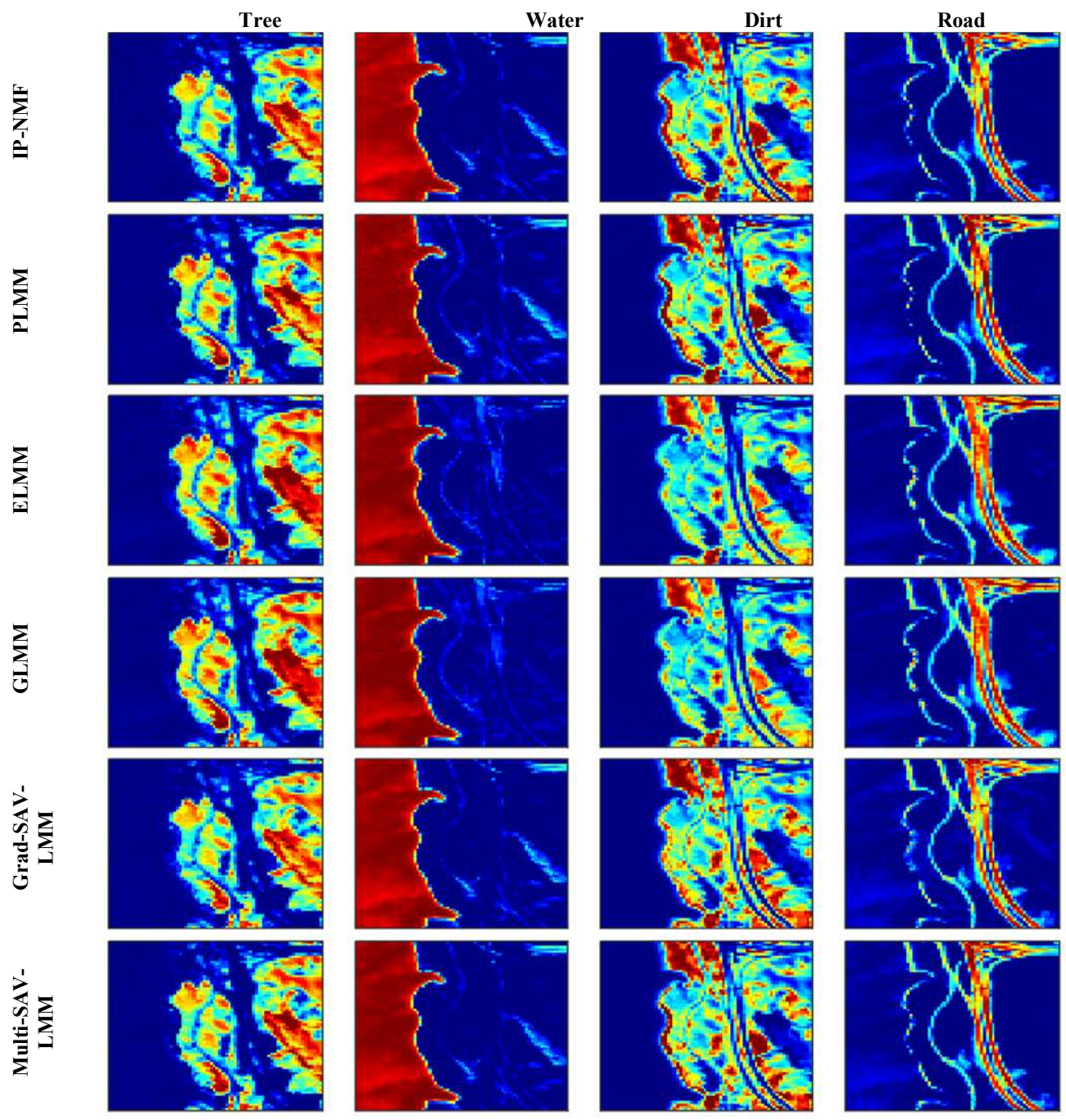

**Fig. 6** Obtained abundance maps for RHI-2.

RHI-2 represents a more rural environment, allowing the evaluation of the proposed methods in a different context with distinct land cover types (water, dirt, tree, and road, as stated above). The proposed approaches achieve good results. The abundance maps estimated by the proposed methods are consistent with the RGB representation of the reference image (Fig. 4). These results confirm the adaptability and effectiveness of the proposed approaches across various real-world scenarios.

## V. Conclusion

In this paper, we have introduced two novel algorithms based on NMF that optimize an additive linear mixing model, namely Grad-SAV-LMM and Multi-SAV-LMM, for hyperspectral spectral unmixing explicitly addressing spectral variability. Extensive experiments conducted on synthetic data and two real datasets show that these methods achieve accurate estimation of abundance maps while obtaining good results in terms of spectral performances. Their ability to model spectral variability leads to more precise and reliable unmixing, especially in complex real-world environments. These methods also demonstrate adaptability across different types of scenes, making them useful for various remote sensing applications such as environmental monitoring and urban planning. Additionally, we introduce a novel initialization process that improves the results of the proposed algorithms and the tested state-of-the-art techniques. The results obtained in this study open up several promising areas of research in the future. Further work could aim to extend the proposed approaches to more complex and realistic scenarios, improve their robustness and computational efficiency, and evaluate their performance under various experimental conditions. Additionally, future studies could explore the integration of advanced optimization or machine learning techniques to improve the accuracy and generalization capabilities of the proposed methods.